%% file: emnlp2022.tex
\pdfoutput=1

\documentclass[11pt]{article}

\usepackage[]{EMNLP2022}
\usepackage{tikz}
\usetikzlibrary{arrows,positioning,calc}

\usepackage{times}
\usepackage{latexsym}
\usepackage{multirow}

\usepackage[T1]{fontenc}

\usepackage[utf8]{inputenc}

\usepackage{microtype}

\usepackage{inconsolata}
\usepackage{subfig}
\usepackage{booktabs}

\input{math_commands.tex}

\DeclareMathOperator{\ffn}{FFN}
\DeclareMathOperator{\gelu}{GELU}

\DeclareMathOperator{\LN}{LN}

\title{CTL++: Evaluating Generalization on Never-Seen Compositional Patterns of Known Functions, and Compatibility of Neural Representations\looseness=-1}

\author{R\'obert Csord\'as$^{1}$ ~ Kazuki Irie$^{1}$ ~ J\"urgen Schmidhuber$^{1,2}$\\
  $^1$The Swiss AI Lab IDSIA, USI \& SUPSI, Lugano, Switzerland \\
  $^2$AI Initiative, KAUST, Thuwal, Saudi Arabia \\
  \texttt{\{robert, kazuki, juergen\}@idsia.ch}
}

\begin{document}
\maketitle
\begin{abstract}

Well-designed diagnostic tasks have played a key role in studying the failure of neural nets (NNs) to generalize systematically.
Famous examples include SCAN and Compositional Table Lookup (CTL).
Here we introduce CTL++, a new diagnostic dataset based on compositions of unary symbolic functions.
While the original CTL is used to test length generalization or \textit{productivity}, CTL++ is designed to test \textit{systematicity} of NNs, that is, their capability to generalize to unseen compositions of known functions.
CTL++ splits functions into groups and tests performance on group elements composed in a way not seen during training.
We show that recent CTL-solving Transformer variants  fail on CTL++.
The simplicity of the task design allows for fine-grained control of task difficulty, as well as many insightful analyses.
For example, we measure how much overlap between groups is needed by tested NNs for learning to compose.
We also visualize how learned symbol representations in outputs of functions from different groups are compatible in case of success but not in case of failure.
These results provide insights into failure cases reported on more complex compositions in the natural language domain. Our code is public.\looseness=-1\footnote{\url{https://github.com/robertcsordas/ctlpp}}

\end{abstract}

\section{Introduction}

Neural networks (NNs) should ideally learn  from training data to generalize \textit{systematically} \citep{fodor1988connectionism}, by learning generally applicable rules instead of pure pattern matching.
Existing NNs, however, typically don't.
For example, in the context of sequence processing NNs,
superficial differences between training and test distributions, e.g., with respect to input sequence length or unseen input/word combinations, are enough to prevent current NNs from generalizing \citep{lake2017generalization}.
Training on a large amounts of data  might alleviate the problem, but it is infeasible to cover all possible lengths and combinations.

Indeed, while large language models trained on a large amounts of data have obtained impressive results \citep{gpt3},
they often fail on tasks requiring simple algorithmic reasoning, e.g., simple arithmetics \citep{rae2021scaling}.
A promising way to achieve systematic generalization is to make NNs more compositional \citep{schmidhuber90composition}, by reflecting and exploiting the hierarchical structure of many problems either within some NN's learned weights, or through tailored NN architectures.
For example, recent work by
\citet{csordas2021ndr} proposes architectural modifications to the standard Transformer \citep{vaswani2017attention} motivated by the principles of compositionality. The resulting Neural Data Router (NDR) exhibits strong \textit{length generalization} or \textit{productivity} on representative datasets such as Compositional Table Lookup (CTL; \citet{liska2018memorize, hupkes2018learning}).

The focus of the present work is on \textit{systematicity}: the capability to generalize to unseen compositions of known functions/words.
That is crucial for learning to process natural language or to reason on algorithmic problems without an excessive amount of training examples.
Some of the existing benchmarks (such as COGS \citep{kim2020cogs} and PCFG \citep{hupkes2019compositionality}) are almost solvable by plain NNs with careful tuning \citep{csordas2021devil}, while others, such as CFQ \citep{keysers2020measuring}, are much harder.
A recent analysis of CFQ by \citet{bogin2022unobserved} suggests that the difficult examples have a common characteristic: they contain some
local structures (describable by parse trees)
which are not present in the training examples.
These findings provide hints for constructing both challenging and intuitive (simple to define and analyze) diagnostic tasks for testing
systematicity.
We propose CTL++, a new diagnostic dataset building upon CTL.
CTL++ is basically as simple as the original CTL in terms of task definition, but adds the core challenge of compositional generalization absent in CTL.
Such simplicity allows for insightful analyses:
one low-level reason for the failure to generalize compositionally appears to be the failure to learn functions whose outputs are symbol representations compatible with inputs of other learned neural functions.
We will visualize this.

Well-designed diagnostic datasets have historically contributed to studies of systematic generalization in NNs. Our CTL++ strives to continue this tradition.

\section{Original CTL}
Our new task (Sec.~\ref{sec:new_task}) is based on the CTL task \citep{liska2018memorize, hupkes2018learning, dubois2020location} whose examples consist of compositions of bijective unary functions defined over a set of symbols.
Each example in the original CTL is defined by one input symbol and a list
of functions to be applied sequentially, i.e., the
first function is applied to the input symbol and the resulting output
becomes the input to the second function, and so forth.
The functions are bijective and randomly generated.
The original CTL uses eight different symbols.
We represent each symbol by a natural number,
and each function by a letter.
For example, `\texttt{d a b 3}' corresponds to the expression $d(a(b(3)))$.
The model has to predict the corresponding output symbol
(this can be viewed as a sequence classification task).
When the train/test distributions are independent and identically distributed (IID), even the basic Transformer achieves perfect test accuracy \citep{csordas2021ndr}.
The task becomes more interesting when test examples are longer than training examples.
In such a \textit{productivity} split, which is the common setting of the original CTL \citep{dubois2020location, csordas2021ndr}, standard Transformers fail, while NDR and bi-directional LSTM work perfectly.

\section{Extensions for Systematicity: CTL++}
\label{sec:new_task}

To introduce a \textit{systematicity} split to the CTL framework, we divide the set of functions into disjoint \textit{groups} and restrict the sampling process such that some patterns of compositions between group elements are never sampled for training, only for testing.
Based on this simple principle, we derive three variations of CTL++.
They differ from each other in terms of
compositional patterns used for testing (excluded from training) as described below.
We'll also visualize the difference using \textit{sampling graphs}
in which the nodes represent the groups, and the edges specify possible compositional patterns.
The colors of the edges reflect when the edges are used: black for both training and testing, {\color{blue} blue}  for training, and {\color{red} red} only for testing.

\paragraph{Variation `A' (as in `Alternating').} 

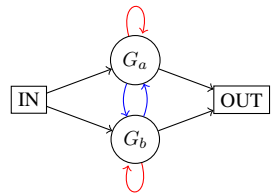
\begin{figure}[t!]
\begin{center}
\begin{tikzpicture}[scale=0.7, every node/.style={transform shape}]
    \node[shape=rectangle,draw=black] (IN) at (0,0) {IN};
    \node[shape=circle,draw=black] (Ga) at (2,0.75) {$G_a$};
    \node[shape=circle,draw=black] (Gb) at (2,-0.75) {$G_b$};
    \node[shape=rectangle,draw=black] (OUT) at (4,0) {OUT};

    \path [->] (IN) edge[black] (Ga);
    \path [->] (IN) edge[black] (Gb);
    \path [->] (Gb) edge[red,loop below] (Gb);
    \path [->] (Ga) edge[red,loop above] (Ga);
    \path [->] (Ga) edge[black] (OUT);
    \path [->] (Gb) edge[black] (OUT);
    \path [->] (Ga) edge[blue, bend right=20] (Gb);
    \path [->] (Gb) edge[blue, bend right=20] (Ga);
\end{tikzpicture}
\caption{Sampling graph for variant `A.'}
\label{fig:sg_gctl_a}
\end{center}
\end{figure}

Here functions are divided in groups $G_a$ and $G_b$. 
During training, successively composed functions are sampled
from different groups in an alternating way---i.e., successive functions cannot be from the same group.
During testing, however, only functions from the same group can be composed. The sampling graph is shown in Fig.~\ref{fig:sg_gctl_a}.
Importantly, the single function applications are part of the training set, to allow the model to learn common input/output symbol representations for the interface between different groups.

\paragraph{Variation `R' (as in `Repeating').} 
This variant is the complement of variation `A' above.
To get a training example, either $G_a$ or $G_b$ is sampled, and all functions in that example are sampled from that same group for the whole sequence.
In test examples, functions are sampled in an alternating way.
There is thus no exchange of information between the groups, except for the shared input embeddings and the output classification weight matrix.
The sampling graph is like in Fig.~\ref{fig:sg_gctl_a} for `A' except that blue edges should become red and vice versa (see Fig.~\ref{fig:sg_gctl_p} in the appendix).

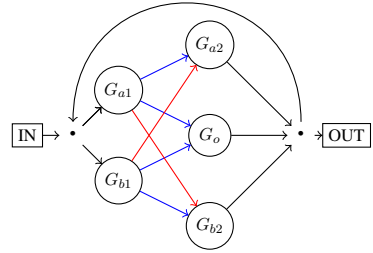
\begin{figure}[t!]
\begin{center}
\vspace{-3mm}
\begin{tikzpicture}[scale=0.6, every node/.style={transform shape}]
    \node[shape=rectangle,draw=black] (IN) at (0,0) {IN};
    \node[shape=circle,draw=white] (fakein) at (1,0) {\textbullet};
    \node[shape=circle,draw=black] (Ga1) at (2,1) {$G_{a1}$};
    \node[shape=circle,draw=black] (Gb1) at (2,-1) {$G_{b1}$};
    \node[shape=circle,draw=black] (Ga2) at (4,2) {$G_{a2}$};
    \node[shape=circle,draw=black] (Go) at (4,0) {$G_{o}$};
    \node[shape=circle,draw=black] (Gb2) at (4,-2) {$G_{b2}$};
    \node[shape=circle,draw=white] (fakeout) at (6,0) {\textbullet};
    \node[shape=rectangle,draw=black] (OUT) at (7,0) {OUT};

   \path [->] (IN) edge[black] (fakein);
   \path [->] (fakein) edge[black] (Ga1);
   \path [->] (fakein) edge[black] (Gb1);
   \path [->] (Ga1) edge[blue] (Ga2);
   \path [->] (Gb1) edge[blue] (Gb2);
   \path [->] (Ga1) edge[blue] (Go);
   \path [->] (Gb1) edge[blue] (Go);
   \path [->] (fakein) edge[black] (Ga1);
   \path [->] (Ga2) edge[black] (fakeout);
   \path [->] (Gb2) edge[black] (fakeout);
   \path [->] (Go) edge[black] (fakeout);
   \path [->] (fakeout) edge[black] (OUT);
   
   \path [->] (Ga1) edge[red] (Gb2);
   \path [->] (Gb1) edge[red] (Ga2);
   
   \path [->] (fakeout) edge[black, bend right=90, min distance=3.5cm] (fakein);
  
\end{tikzpicture}
\caption{Sampling graph for Variant `S'}
\label{fig:sg_gctl_p2}
\end{center}
\end{figure}

\paragraph{Variation `S' (as in `Staged').}
In this variant, functions are divided into five disjoint groups: $G_{a1}$, $G_{a2}$, $G_{b1}$, $G_{b2}$ and $G_{o}$.
As indicated by the indices, each group belongs to one of the two \textit{paths} (`a' or `b') and one of the two \textit{stages} (`1' or `2'), except for $G_o$ which only belongs to stage `2' shared between paths `a' and `b' during training.
The corresponding sampling graph is shown in Fig.~\ref{fig:sg_gctl_p2}.
To get a training example, 
we sample an integer $K$ which defines the sequence length as $2K+1$, and iterate
the following process for $k \in [0, .., K]$ and $i = 2k$:
we first sample a path $p \in \{ a, b \}$ and then a function $f_i$ from $G_{p1}$ and a function $f_{i+1}$ from $G_{p2} \cup G_{o}$.
Each example always contains an even number of functions, and no isolated single function application is part of training, unlike in the previous two variants.
For testing, we sample a path $p \in \{ a, b \}$ and a function $f_i$ from $G_{p1}$, but then sample a function $f_{i+1}$ from $G_{\{a,b\} \setminus \{p\}2}$, which results in a compositional pattern never seen during training.

The unique feature of this variant is the use of two stages: as can be seen in Fig.~\ref{fig:sg_gctl_p2}, during training,
given a path $p \in \{ a, b \}$, outputs of any functions belonging to $G_{p1}$ are only observed by the functions belonging to $G_{p2}$, i.e., the stage `2' group belonging to the same path $p$, or $G_o$.
Hence, if $G_o = \emptyset$, the model has no incentive to learn common representations for the interface between $G_{a1}$ and $G_{b1}$:
to solve the training examples, it suffices to learn
output representations of $G_{a1}$ which are `compatible' with the input representations of $G_{a2}$; similarly for $G_{b1}$ and $G_{b2}$.
There is no reason for outputs of $G_{a1}$ to be compatible with the inputs of $G_{b2}$ (analogously for $G_{b1}$ and $G_{a2}$) which is required at test time.
The size of $G_o$ is our first parameter
for controlling task difficulty (the y-axis of Fig.~\ref{fig:ctl_p2} which we will present later).

We introduce further restrictions: for each function $f \in G_o$, we define a set of symbols $S^f_{a}$ for $G_{a1}$ (and $S^f_{b}$ for $G_{b1}$), and we only allow for sampling $f$ if the output symbol of function from $G_{a1}$ (or $G_{b1}$) belongs to $S^f_{a}$ (or $S^f_{b}$).
This allows for defining another control parameter:
the number of overlapping symbols between $S^f_{a}$ and $S^f_{b}$ (same for all $f$; the x-axis of Fig.~\ref{fig:ctl_p2}).
Note that we ensure that the union of shared symbols defined for functions in $G_o$ cover all possible symbols.
This might not be the case in a more realistic scenario, but as we'll see, the standard models already struggle in this setting.
By controlling these two parameters, we precisely
control the degree of overlap offered by $G_o$ in terms of both the number of functions and symbols.
Ideal models should be ``sample efficient'' in terms of this overlap, since we cannot expect the training set to contain all combinations of such overlaps in a practical scenario with semantically rich domains such as natural language.

\section{Results}

We evaluate standard CTL-tested models on the new CTL++ task, including: 
the Transformer \citep{vaswani2017attention} with shared layers \citep{dehghani2019universal}, the neural data router (NDR) \citep{csordas2021ndr}, and the bi-directional LSTM \citep{hochreiter1997long, schuster1997bidirectional,graves:icann05}.
Recall that both NDR and bi-directional LSTM are reported to perfectly solve the original CTL's length generalization split \citep{csordas2021ndr}, unlike the Transformer.
Further experimental details can be found in Appendix \ref{app:exp}.

\begin{table}[h]
    \centering
    \small
    \caption{Results on the task variants `A' and `R.' Mean and standard deviation are computed using 25 seeds.}
    \label{tab:single_stage_res}
    \begin{tabular}{l l c c }
        \toprule
\multirow{2.5}{*}{Model} & \multirow{2.5}{*}{Dataset} &  \multicolumn{2}{c}{Accuracy} \\ \cmidrule{3-4}
 & & IID  & OOD  \\
\midrule
\multirow{2}{*}{Bi-LSTM} & A & $1.00 \pm 0.00$ & $0.95 \pm 0.03$ \\
 & R & $1.00 \pm 0.00$ & $1.00 \pm 0.00$ \\
\midrule
\multirow{2}{*}{Transformer} & A & $1.00 \pm 0.00$ & $0.21 \pm 0.09$ \\
 & R & $1.00 \pm 0.00$ & $0.75 \pm 0.25$ \\
\midrule
\multirow{2}{*}{NDR} & A & $1.00 \pm 0.00$ & $0.34 \pm 0.26$ \\
 & R & $1.00 \pm 0.01$ & $0.75 \pm 0.27$ \\
        \bottomrule
    \end{tabular}
\end{table}

\subsection{Results on Variants `A' and `R'}
\label{sec:a_r}
Table \ref{tab:single_stage_res} shows the performance overview for `A' and `R.'
The OOD (out-of-distribution) column indicates the train/test data sampling processes described above.
As a reference, we also report the IID cases where the training example sampling graph is also used for testing.
Our initial expectation was that the pressure from shared input/output embeddings is sufficient for these models to learn common symbol representations for all functions.
However, we observe that only the bi-LSTM solves these tasks consistently across seeds.
Interestingly, the NDR, which perfectly performs on the length generalization split of CTL and beyond \citep{csordas2021ndr}, performs poorly on both the `A' and `R' variants of CTL++.
Tested with 25 seeds, only 32\% of the seeds (out of 25) achieve over 95\% accuracy for NDR on `R' (20\% for the standard Transformer).
The success rate is 0\% for the `A' variant\footnote{`A' turns out to be harder than `R.' We speculate that 
in both `A' and `R', given that the training set contains single function applications with shared input/ouput embeddings, the learned symbol representation of all functions should be compatible with each other to some extent, but with some ``deviation'' from perfect compatibility in case of failure.
Such ``deviations'' might accumulate in case of `A' where we sample all functions in each sequence from a single group at test time.}.
These simple tasks thus turn out to be good first diagnostic tasks for testing systematicity.

\begin{figure}
\begin{center}
\subfloat[Example for symbol `6'. Perfect clustering w.r.t.~the function groups is observed.]{\includegraphics[width=0.45\columnwidth]{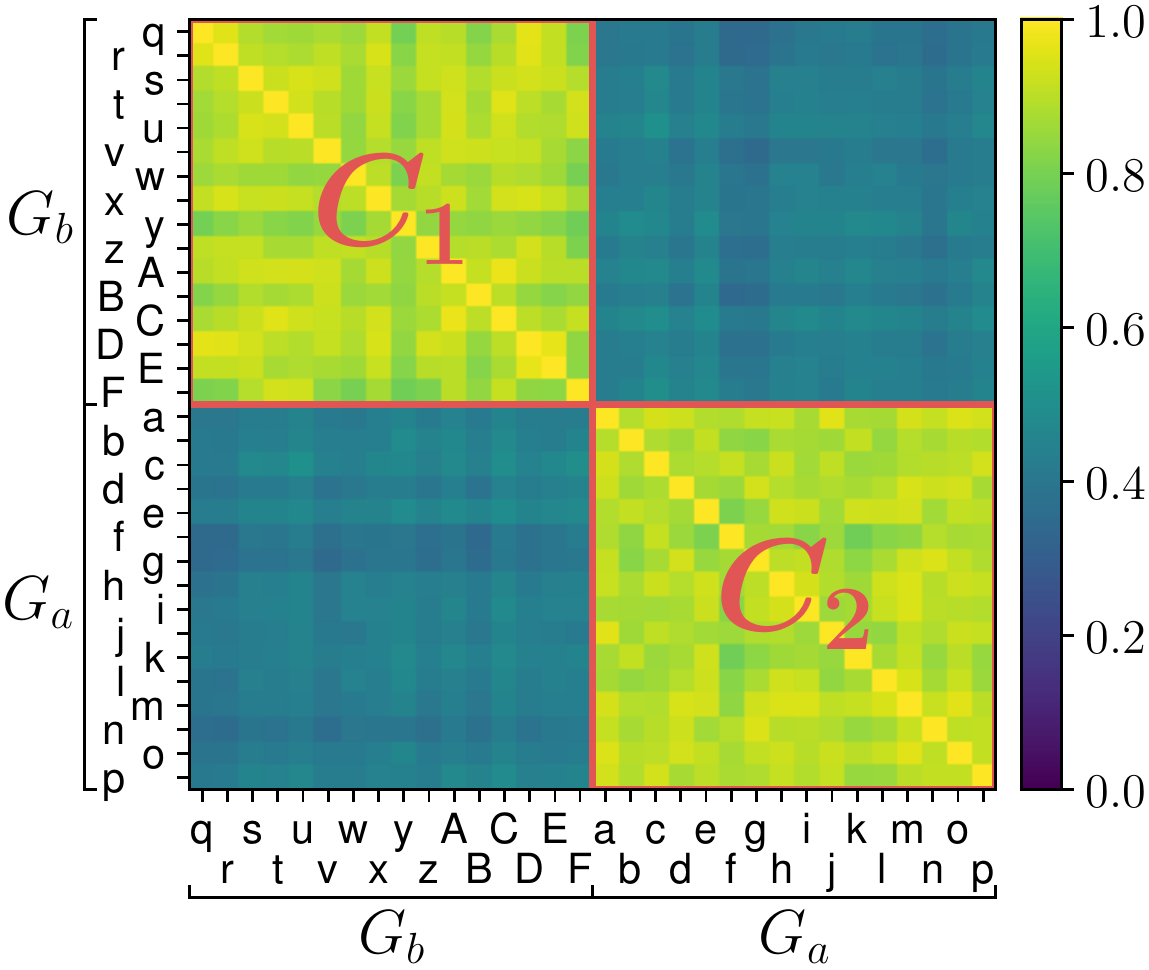}\label{fig:bad_total}}
\qquad
\subfloat[Example for symbol `3'. Partial clustering w.r.t.~the function groups is observed.]{\includegraphics[width=0.45\columnwidth]{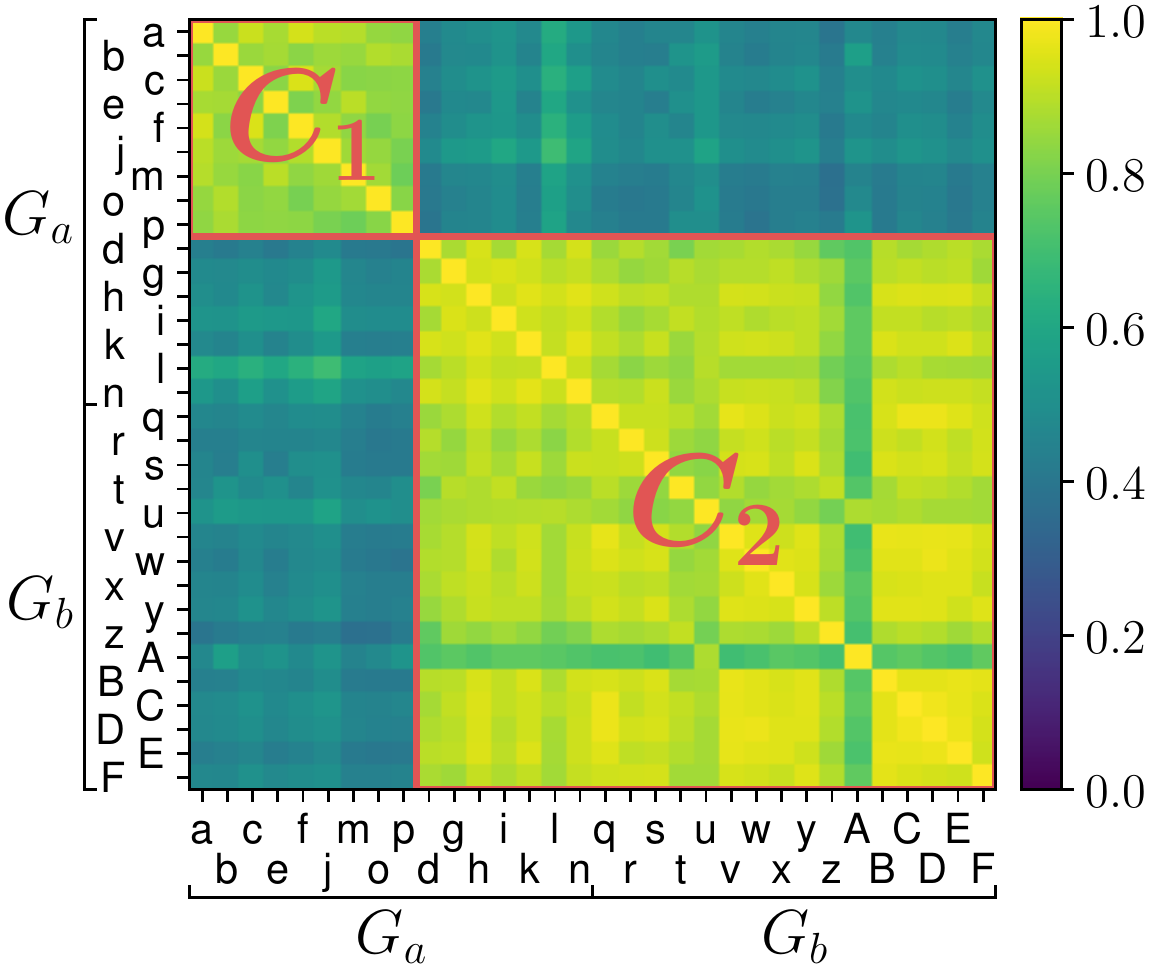}\label{fig:bad_partial}}
\caption{Cosine similarity of output representations of different functions representing the same symbol for the NDR with a seed that fails on `R.'
}
\label{fig:repr_bad}
\end{center}
\end{figure}

\paragraph{Analysis.}
The small input/output space of this task allows for an exhaustive analysis of the learned symbol representations.
Specifically, given an \textit{output} symbol $s'$, for each function $f \in G_a \cup G_b$, we can find a unique input symbol $s$ such that $s' = f(s)$ (because all functions defined for this task are bijective).
Hence, for a fixed symbol $s'$, for all functions $f$, we can extract the learned vector representation of this symbol $s'$ at the output of  $f$ as the vector of the layer beneath the final classification layer when we feed `$f$ $s$' to the network.
Then we can compare the extracted representations (of a fixed symbol for different functions) by computing their cosine similarities.

Here we compare representations learned with successful/failed seeds for NDR in variation `R.'
Fig.~\ref{fig:bad_total} and \ref{fig:bad_partial} show the results for two different (output) symbols `6' and `3' from the same \textit{failed} seed.
In both cases we observe two clusters ($C_1$ and $C_2$):
two separate/different representations are learned for the same symbol (by abuse of notation, we also refer to the corresponding representations as $C_1$ and $C_2$).
In the case of symbol `6' in Fig.~\ref{fig:bad_total}, we observe perfect/strict clustering in line with the group of the applied function;
$C_1$ and $C_2$ are representations of symbol `6` learned by functions belonging to $G_a$ and $G_b$ respectively.
This is problematic since functions in $G_a$ never see symbol `6' represented as $C_2$ during training (analogously for functions in $G_b$ with representation $C_1$).
As a consequence, during testing, when a function $f_a \in G_a$
is applied after a function $f_b \in G_b$,
$f_b$ may output symbol `6' represented as $C_2$, and pass it to $f_a$, but in principle, $f_a$ can not ``understand/interpret'' $C_2$ as representing symbol `6.'
This naturally prevents cross-group generalization.
In the case of symbol `3' shown in Fig.~\ref{fig:bad_partial}, some of the functions yield the same symbol representations as certain functions from the other group (see the cluster $C_2$ at the lower right: a good trend),
but we still have a small cluster ($C_1$ at the upper left)  consistent only among elements of $G_a$.
Hence, cross-group generalization can still fail because the functions in $G_b$ never see symbol `3' represented as $C_1$ during training but only during testing.
In contrast, for \textit{successful} seeds,
we do not observe any of these clusters for any symbols (see Fig.~\ref{fig:all_cossim_good} in the appendix).
A single representation shared across all functions is learned for each symbol.
Further quantitative analysis can be found in Appendix \ref{app:more_analysis}.

\begin{figure}[t]
\begin{center}
\includegraphics[width=60mm]{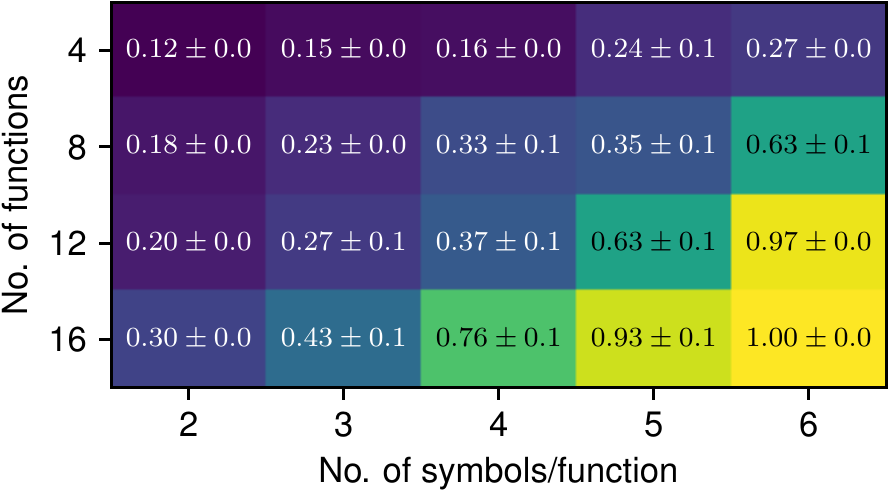}

\caption{Test accuracy of NDR on the `S' variant. The total number of symbols is 8, and the number of functions is 32. The y-axis shows the number of overlapping functions, while the x-axis shows the number of symbols shared between two groups for each function in $G_o$ during training (Sec.~\ref{sec:new_task}).
Results for the Transformer and LSTM 
are reported in the appendix (Figs.~\ref{fig:ctl_p2_trafo} and \ref{fig:ctl_p2_rnn}).
}
\label{fig:ctl_p2}
\end{center}
\end{figure}

\subsection{Results of Staged Variant `S'}
As described in Sec.~\ref{sec:new_task}, variant `S' is designed to evaluate models at different task difficulty levels determined by the number of overlapping functions and symbols during training.
Fig.~\ref{fig:ctl_p2} shows the corresponding performance overview for  NDR.
The overall picture is similar for bi-LSTM and Transformer (see Figs.~\ref{fig:ctl_p2_trafo} and \ref{fig:ctl_p2_rnn} in the appendix). 
We observe that to achieve 100\% accuracy, half of the possible functions should overlap (16/32), as well as most of the possible symbols seen for each function (6/8).
This implies an unrealistically large amount of data for real world scenarios, where the ``functions'' might correspond to more complex operations with multiple input  arguments (as in the CFQ case).
This calls for developing approaches that achieve higher accuracy in the upper left part of Fig.~\ref{fig:ctl_p2}.

\section{Conclusion}

Motivated by the historically crucial role of diagnostic datasets for research on systematic generalization of NNs, we propose a new dataset called CTL++.
Unlike the classic CTL dataset, typically used for testing productivity,  CTL++ is designed for testing systematicity.
We propose three variants, `A,' `R,' and `S.'
Despite their simplicity,
even the CTL-solving Transformer variant  fails on `A' and `R.'
Using `S,' we show that existing approaches require impractically large amounts of examples to achieve perfect compositional generalization.
The small task size  allows for conducting exhaustive visualizations of (in)compatibility of learned symbol representations in outputs of functions with inputs of subsequent functions.
Of course, the ultimate goal is to go beyond just solving CTL++. 
Nevertheless, 
we hope CTL++ will become one of the standard diagnostic datasets for testing systematicity of NNs.

\section*{Limitations}
Achieving 100\% on this dataset may be a necessary condition
for NNs capable of systematic generalization,
but certainly not a sufficient one.
In practice, there may be many
reasons which prevent
NNs from generalizing systematically in other tasks or more generally on real world data.
Compare the original CTL dataset for evaluating productivity:
 \citet{csordas2021ndr} show that some models that achieve
100\% on CTL still fail in other tasks such as ListOps.
This is why we refer to CTL++ as a simple \textit{diagnostic} dataset for
testing systematicity of NNs. Nevertheless, it allows for uncovering certain important failure modes of NNs.

\section*{Acknowledgments}
This research was partially funded by ERC Advanced grant no: 742870, project AlgoRNN,
and by Swiss National Science Foundation grant no: 200021\_192356, project NEUSYM.
We are thankful for hardware donations from NVIDIA and IBM.
The resources used for this work were partially provided by Swiss National Supercomputing Centre (CSCS) project s1154.

\bibliography{anthology,custom}
\bibliographystyle{acl_natbib}

\clearpage
\appendix

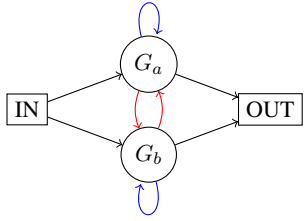
\begin{figure}[t!]
\begin{center}
\begin{tikzpicture}[scale=0.8, every node/.style={transform shape}]
    \node[shape=rectangle,draw=black] (IN) at (0,0) {IN};
    \node[shape=circle,draw=black] (Ga) at (2,0.75) {$G_a$};
    \node[shape=circle,draw=black] (Gb) at (2,-0.75) {$G_b$};
    \node[shape=rectangle,draw=black] (OUT) at (4,0) {OUT};

    \path [->] (IN) edge[black] (Ga);
    \path [->] (IN) edge[black] (Gb);
    \path [->] (Gb) edge[blue,loop below] (Gb);
    \path [->] (Ga) edge[blue,loop above] (Ga);
    \path [->] (Ga) edge[black] (OUT);
    \path [->] (Gb) edge[black] (OUT);
    \path [->] (Ga) edge[red, bend right=20] (Gb);
    \path [->] (Gb) edge[red, bend right=20] (Ga);
\end{tikzpicture}
\caption{Sampling graph for variant `R.'}
\label{fig:sg_gctl_p}
\end{center}
\end{figure}

\section{Experimental Details}
\label{app:exp}

\subsection{Modified NDR architecture}
\label{app:ndr_modification}

We experimentally found some architectural modifications to the original NDR \citep{csordas2021ndr} that yield faster convergence and produce more stable results than the original architecture.
Here we describe our modifications.
We use the GELU activation function \citep{hendrycks2016gaussian} instead of ReLU, and a residual connection in the feedforward data path.
Concretely, Eq.~5 in \citet{csordas2021ndr} is replaced by Eq.~\ref{eq:ffn} below, while Eq. 2 is replaced by Eq.~\ref{eq:data} below. We do not use any dropout in Eq.~\ref{eq:ffn}.

\begin{align}
\label{eq:ffn}
\ffn (\vx) &= \mW_2 \gelu(\mW_1 \vx +\vb_1)+\vb_2 \\
\label{eq:data}
\vu^{(i, t+1)} &= \LN(\ffn^\textrm{data}(\va^{(i, t+1)})+ \va^{(i, t+1)})
\end{align}
where $\LN$ denotes layer normalization \citep{ba2016layer}.

\subsection{Hyperparameters}

\paragraph{Dataset.} The train set in all of our experiments consists of 300k examples. The maximum number of composed functions is 6.
We make sure that we obtain an equal number of samples for different lengths whenever possible (in some cases this is impossible because by construction, there are fewer short examples than long ones).
In `A' and `R' variants, single functions are always part of the training set with all possible symbols. There are in total 8 symbols and 32 functions. All samples are presented in a right-to-left manner (e.g. ``c b a 3'').
The IID and OOD test sets contain 1000 examples in all cases.
Our code generates the data for given dataset specifications (number of functions etc). The seed for data generation is fixed.

\begin{figure}[h]
\begin{center}
\includegraphics[width=70mm]{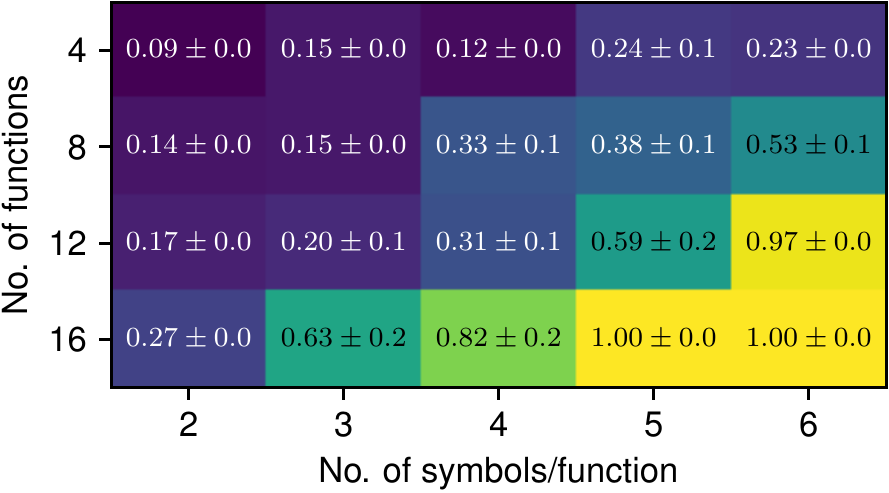}

\caption{Final test performance of the Transformer on variant `S.' The behavior is similar to the one shown in Fig.~\ref{fig:ctl_p2}. For lower numbers of shared functions, performance is worse. Interestingly, however, with 16 shared functions, it outperforms NDR.}
\label{fig:ctl_p2_trafo}
\end{center}
\end{figure}

\begin{figure}[h]
\begin{center}
\includegraphics[width=70mm]{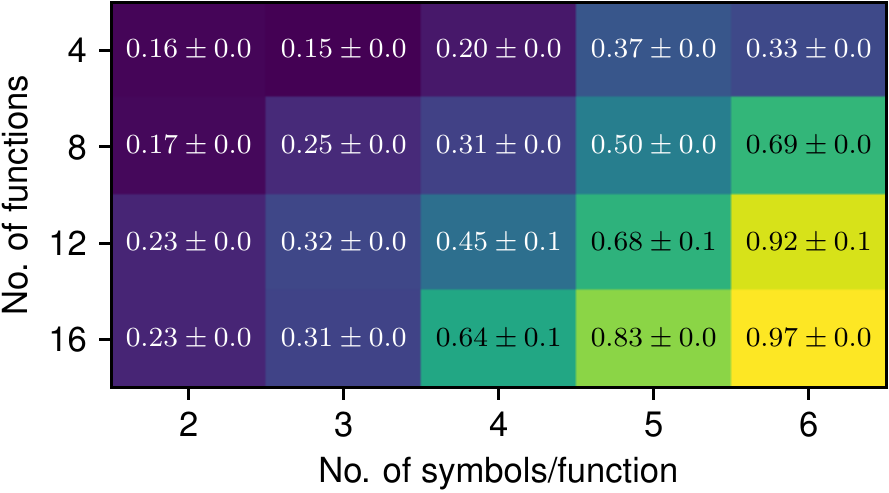}

\caption{Final test performance of bi-directional LSTM on variant `S.' The behavior is similar to the one shown in Fig.~\ref{fig:ctl_p2}. For lower numbers of shared functions, the performance is worse. Interestingly, however,  with 16 shared functions, it significantly underperforms NDR.}
\label{fig:ctl_p2_rnn}
\end{center}
\end{figure}

\paragraph{Training.} Unless noted otherwise, for all of our models, we use a batch size of 512, a learning rate of 0.00015, and a dropout rate \citep{hanson1990stochastic, srivastava2014dropout} of 0.5. We also use a linear learning rate warmup for the first 500 iterations. We use PyTorch's \citep{paszke2019pytorch} adaptive mixed precision and bin the batches by length for greater efficiency. We use the AdamW optimizer \citep{loshchilov2019decoupled}. The NDR and bi-LSTM is trained for 80k and the Transformers for 300k iterations.
We find the standard Transformer to be very unstable even in the IID setting.
In fact, for Table \ref{tab:single_stage_res}, unlike other models trained for 25 seeds, we train the Transformer for 50 seeds:
the 25 seeds used to report mean and std in Table \ref{tab:single_stage_res} are those among 50 which converged within 300k training iterations.
For Figs.~\ref{fig:ctl_p2}, \ref{fig:ctl_p2_trafo} and \ref{fig:ctl_p2_rnn}, five seeds were used for each configuration.
All of our models are trained on a single P100 GPU. The corresponding number of parameters, training steps and average wall-clock time is shown in Table \ref{tab:training_details}.

\begin{table}[h]
    \centering
    \small
    \caption{Training details for different models. Runtime is in hour:min.}
    \label{tab:training_details}
    \begin{tabular}{l r r r }
        \toprule

Model & Num. params & Num. steps  & Runtime  \\
\midrule
Bi-LSTM & 408k & 80k & 0:18 \\
Transformer & 672k & 300k & 3:37 \\
NDR & 679k & 80k & 1:22 \\
        \bottomrule
    \end{tabular}
\end{table}

\paragraph{Models.} For Transformer and NDR, we use 8 layers, and 4 heads.  NDR uses a gate dropout of 0.1, state size of 256, feedforward size of 1024. Transformers use a state size of 128 and feedforward size of 512, layer sharing \citep{dehghani2019universal}, and Transformer-XL-style \citep{dai2019transformer} relative positional encoding. For  bidirectional LSTM, we use 1 layer with 256 units (128 per direction). The gradient is clipped to max norm of 1 for NDR and 5 for Transformer and LSTM. Transformers use a weight decay of $0.0025$.

\section{More Analyses and Plots}
\label{app:more_analysis}
\subsection{Quantitative Analysis of Incompatibility}
\begin{figure}[h]
\begin{center}
\subfloat[Performance of each cluster vs. each group]{\includegraphics[width=0.4\columnwidth]{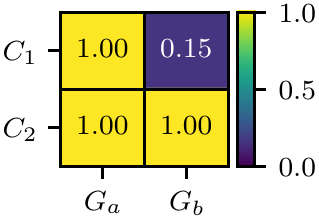}\label{fig:e1}}
\qquad
\subfloat[Performance for each pair of groups]{\includegraphics[width=0.4\columnwidth]{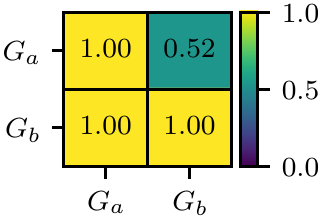}\label{fig:e2}}
\caption{Accuracy measured after two successive function applications, for the symbol corresponding to Fig.~\ref{fig:bad_partial}. (a) shows the proportion of correct outputs when the first function is taken from a given cluster (y-axis), and the second from a given group (x-axis). Clusters are shown in the main diagonal of Fig.~\ref{fig:bad_partial}. (b) is analogous to (a) but using groups as clusters.}
\label{fig:bad_error_analysis}
\end{center}
\end{figure}

Here we provide additional results on the analysis of Sec.~\ref{sec:a_r} conducted for variant `R.'
To quantify the correlation between the clusters ($C_1$ and $C_2$) identified in Fig.~\ref{fig:bad_partial} and the compatibility of representations,
we measured the proportion of correct output classifications, by taking the first function from a given cluster and the second one from a given group, for all pairs of functions for each pair of the form (cluster, group).
The results are shown in Fig.~\ref{fig:bad_error_analysis}.
Fig.~\ref{fig:e1} shows that the first cluster $C_1$ is effectively compatible only with functions from $G_a$, while the second one $C_2$ works with both $G_a$ and $G_b$, as predicted by Fig.~\ref{fig:bad_partial}.
Fig.~\ref{fig:e2} shows the same analysis, but uses the groups to define the cluster.
As predicted by Fig.~\ref{fig:bad_partial}, only roughly half of the functions from $G_a$ generate representations compatible with $G_b$, while all representations generated by functions in $G_b$ are compatible with all in $G_a$.

\subsection{Representative Cosine Similarities}
\label{sec:representative_cossim}

\begin{figure*}[ht]
\begin{center}
\includegraphics[width=1.9\columnwidth]{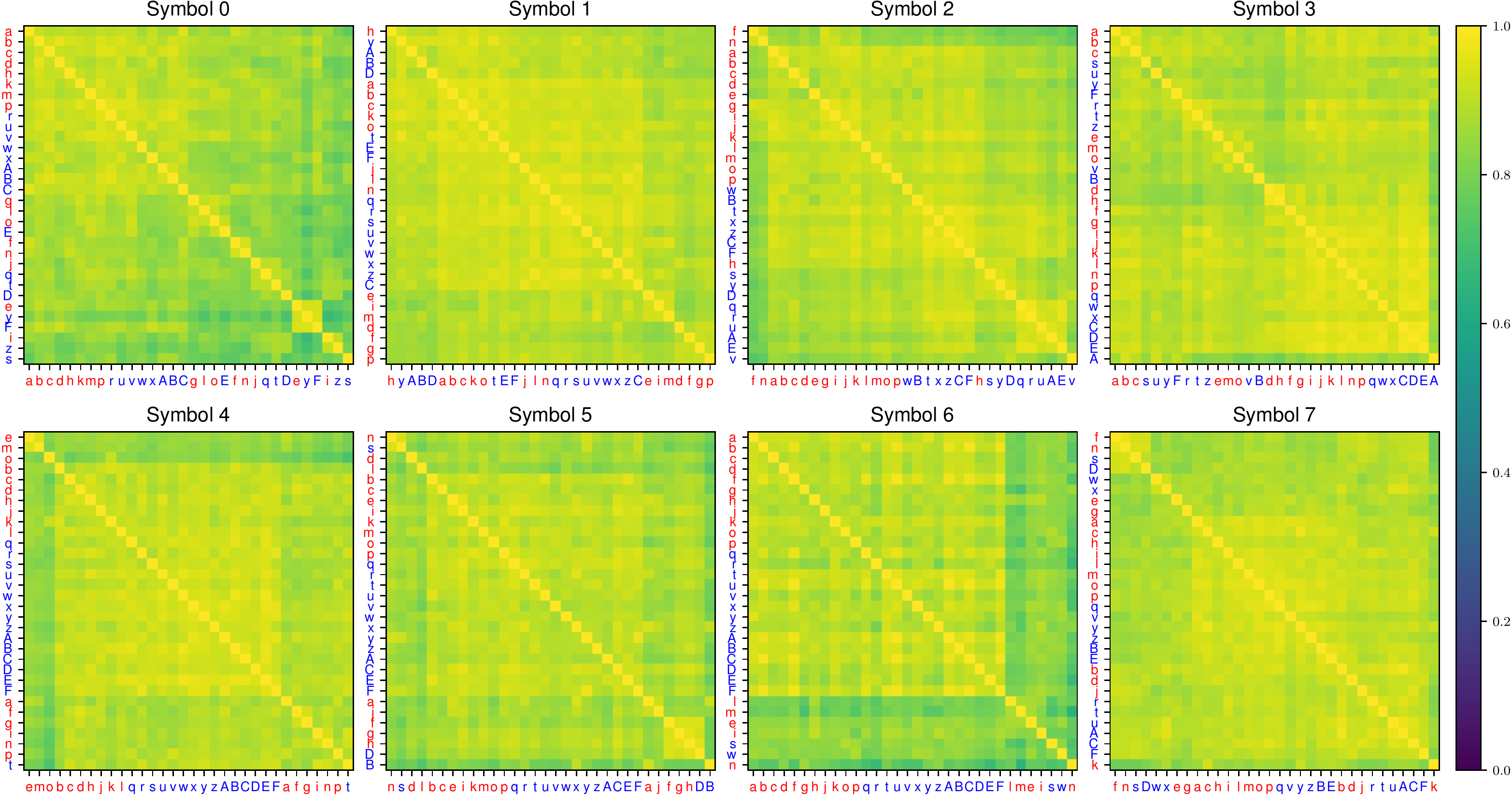}

\caption{Symbol cosine similarity between different functions for NDR on variant `R.' A representative example from a seed that performs \textbf{perfectly} on unseen compositions. Functions indicated by {\color{red}red} belong to $G_a$, by {\color{blue}blue} to $G_b$.} 
\label{fig:all_cossim_good}
\end{center}
\end{figure*}

\begin{figure*}[ht]
\begin{center}
\includegraphics[width=1.9\columnwidth]{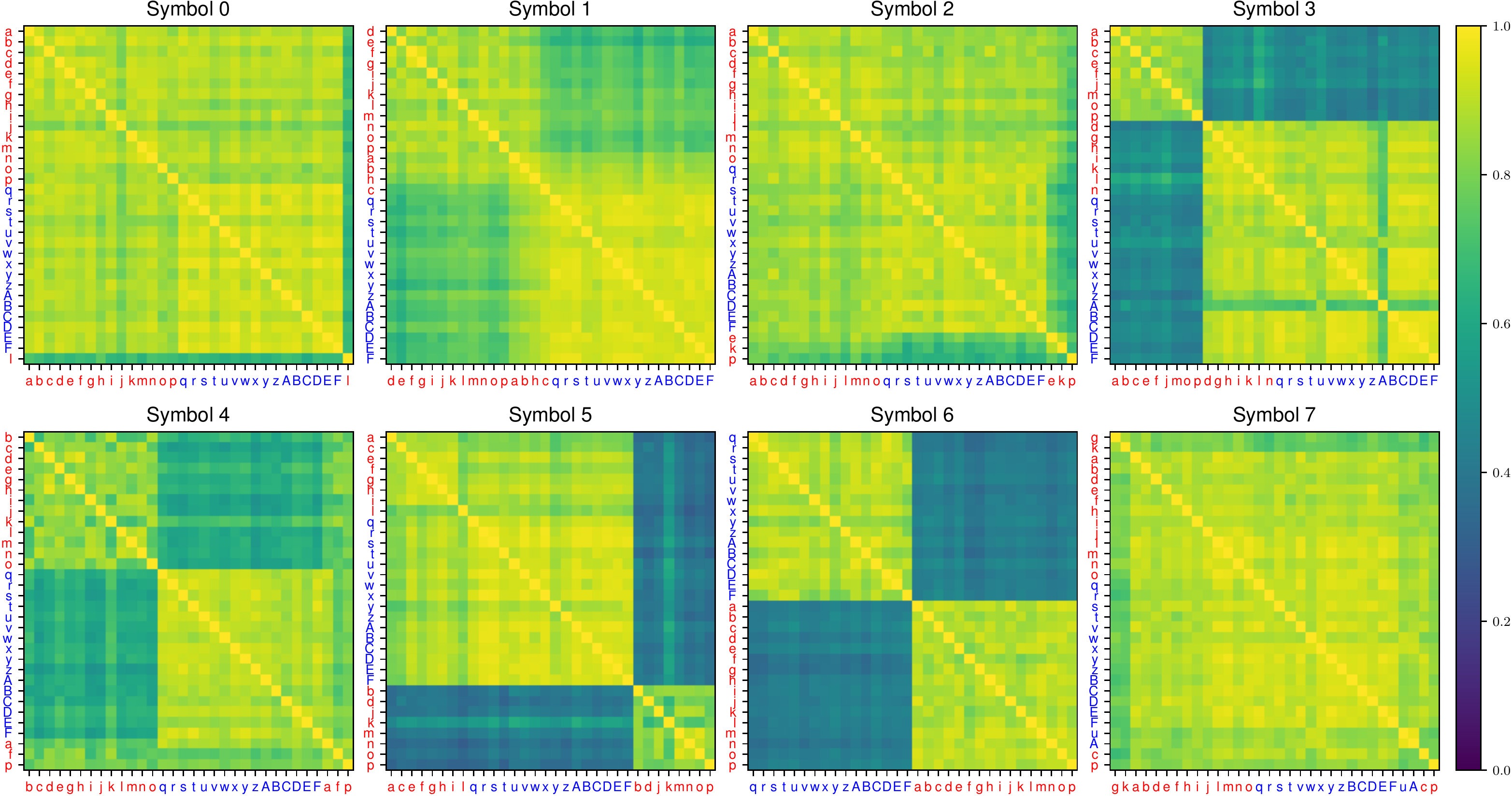}

\caption{Symbol cosine similarity between different functions for NDR  on variant `R.' A representative example from a seed that performs \textbf{poorly} on unseen compositions. Functions indicated by {\color{red}red} belong to $G_a$, by {\color{blue}blue} to $G_b$.}
\label{fig:all_cossim_bad}
\end{center}
\end{figure*}

\begin{figure*}[ht]
\begin{center}
\includegraphics[width=1.9\columnwidth]{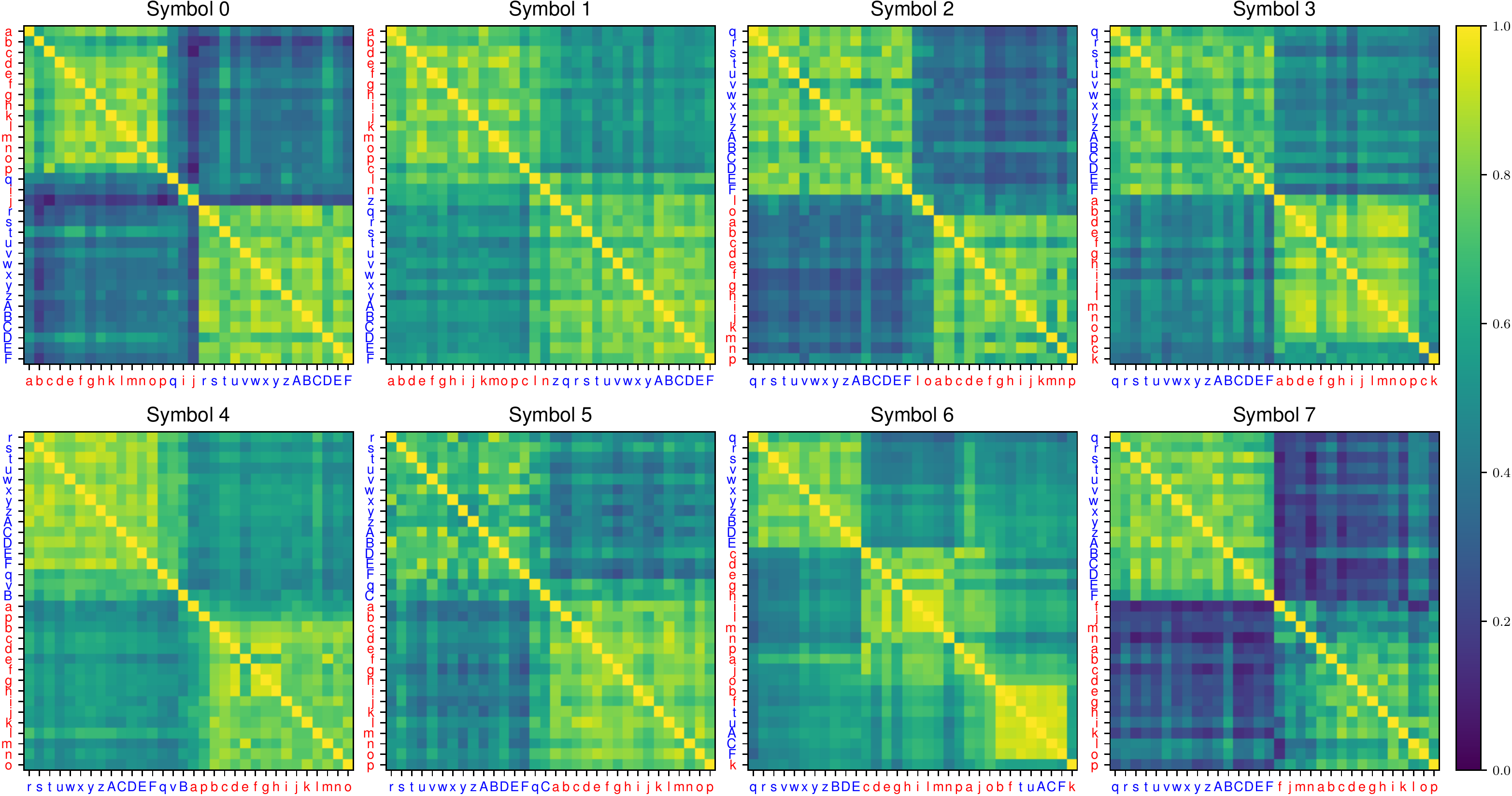}

\caption{Symbol cosine similarity between different functions for NDR on variant `A.' A representative example from a seed that performs \textbf{poorly} on unseen compositions. Functions indicated by {\color{red}red} belong to $G_a$, by {\color{blue}blue} to $G_b$.}
\label{fig:all_cossim_bad_alt}
\end{center}
\end{figure*}

Here we show additional visualizations similar to those of Fig.~\ref{fig:repr_bad}.
In Fig.~\ref{fig:all_cossim_good} and \ref{fig:all_cossim_bad}, we plot cosine similarities of functional outputs for all possible symbols for successful and failed seeds of NDR on variant `R.'
The symbol representations are taken from  the layer right below the final classification layer. They are representative examples; the observation holds over all seeds we inspected. Fig. \ref{fig:all_cossim_bad_alt} shows a similar example for variant `A.'

\vfill\null
\pagebreak

\end{document}

%% file: math_commands.tex
\usepackage{amsmath,amsfonts,bm}

\def\eqref#1{equation~\ref{#1}}

\def\1{\bm{1}}

\def\va{{\bm{a}}}
\def\vb{{\bm{b}}}

\def\vu{{\bm{u}}}

\def\vx{{\bm{x}}}

\def\mW{{\bm{W}}}

\DeclareMathAlphabet{\mathsfit}{\encodingdefault}{\sfdefault}{m}{sl}
\SetMathAlphabet{\mathsfit}{bold}{\encodingdefault}{\sfdefault}{bx}{n}

